\documentclass[sigconf]{acmart}

\usepackage{multirow}
\usepackage{dsfont}
\usepackage{amsfonts}
\usepackage{amsmath}
\usepackage{balance}

\graphicspath{ {figures/} }

\def\BibTeX{{\rm B\kern-.05em{\sc i\kern-.025em b}\kern-.08emT\kern-.1667em\lower.7ex\hbox{E}\kern-.125emX}}

\copyrightyear{2019}
\acmYear{2019}
\acmConference[MM '19]{Proceedings of the 27th ACM International Conference on Multimedia}{October 21--25, 2019}{Nice, France}
\acmBooktitle{Proceedings of the 27th ACM International Conference on Multimedia (MM '19), October 21--25, 2019, Nice, France}
\acmPrice{15.00}
\acmDOI{10.1145/3343031.3351062}
\acmISBN{978-1-4503-6889-6/19/10}

\begin{document}

\fancyhead{}

\title{PDANet: Polarity-consistent Deep Attention Network \\for Fine-grained Visual Emotion Regression}

\author{Sicheng Zhao}
\email{schzhao@gmail.com}
\affiliation{%
  \institution{University of California, Berkeley}
}

\author{Zizhou Jia}
\email{jiazizhou@126.com}
\affiliation{%
  \institution{Tsinghua University}
}

\author{Hui Chen}
\email{jichenhui2012@gmail.com}
\affiliation{%
  \institution{Tsinghua University}
}

\author{Leida Li}
\authornote{Corresponding authors: Leida Li, Guiguang Ding.}
\email{reader1104@hotmail.com}
\affiliation{%
  \institution{Xidian University}
}

\author{Guiguang Ding}
\authornotemark[1]
\email{dinggg@tsinghua.edu.cn}
\affiliation{%
  \institution{Tsinghua University}
}

\author{Kurt Keutzer}
\email{keutzer@berkeley.edu}
\affiliation{%
  \institution{University of California, Berkeley}
}

%
\renewcommand{\shortauthors}{Zhao, et al.}

%
\begin{abstract}
Existing methods on visual emotion analysis mainly focus on coarse-grained emotion classification, \textit{i.e.} assigning an image with a dominant discrete emotion category. However, these methods cannot well reflect the complexity and subtlety of emotions. In this paper, we study the fine-grained regression problem of visual emotions based on convolutional neural networks (CNNs). Specifically, we develop a Polarity-consistent Deep Attention Network (PDANet), a novel network architecture that integrates attention into a CNN with an emotion polarity constraint. First, we propose to incorporate both spatial and channel-wise attentions into a CNN for visual emotion regression, which jointly considers the local spatial connectivity patterns along each channel and the interdependency between different channels. Second, we design a novel regression loss, \textit{i.e.} polarity-consistent regression (PCR) loss, based on the weakly supervised emotion polarity to guide the attention generation. By optimizing the PCR loss, PDANet can generate a polarity preserved attention map and thus improve the emotion regression performance. Extensive experiments are conducted on the IAPS, NAPS, and EMOTIC datasets, and the results demonstrate that the proposed PDANet outperforms the state-of-the-art approaches by a large margin for fine-grained visual emotion regression. Our source code is released at: \url{https://github.com/ZizhouJia/PDANet}.
\end{abstract}

%
%


%

%
\maketitle

\section{Introduction}
\label{sec:Introduction}

\begin{figure}[!t]
\begin{center}
\centering \includegraphics[width=1.0\linewidth]{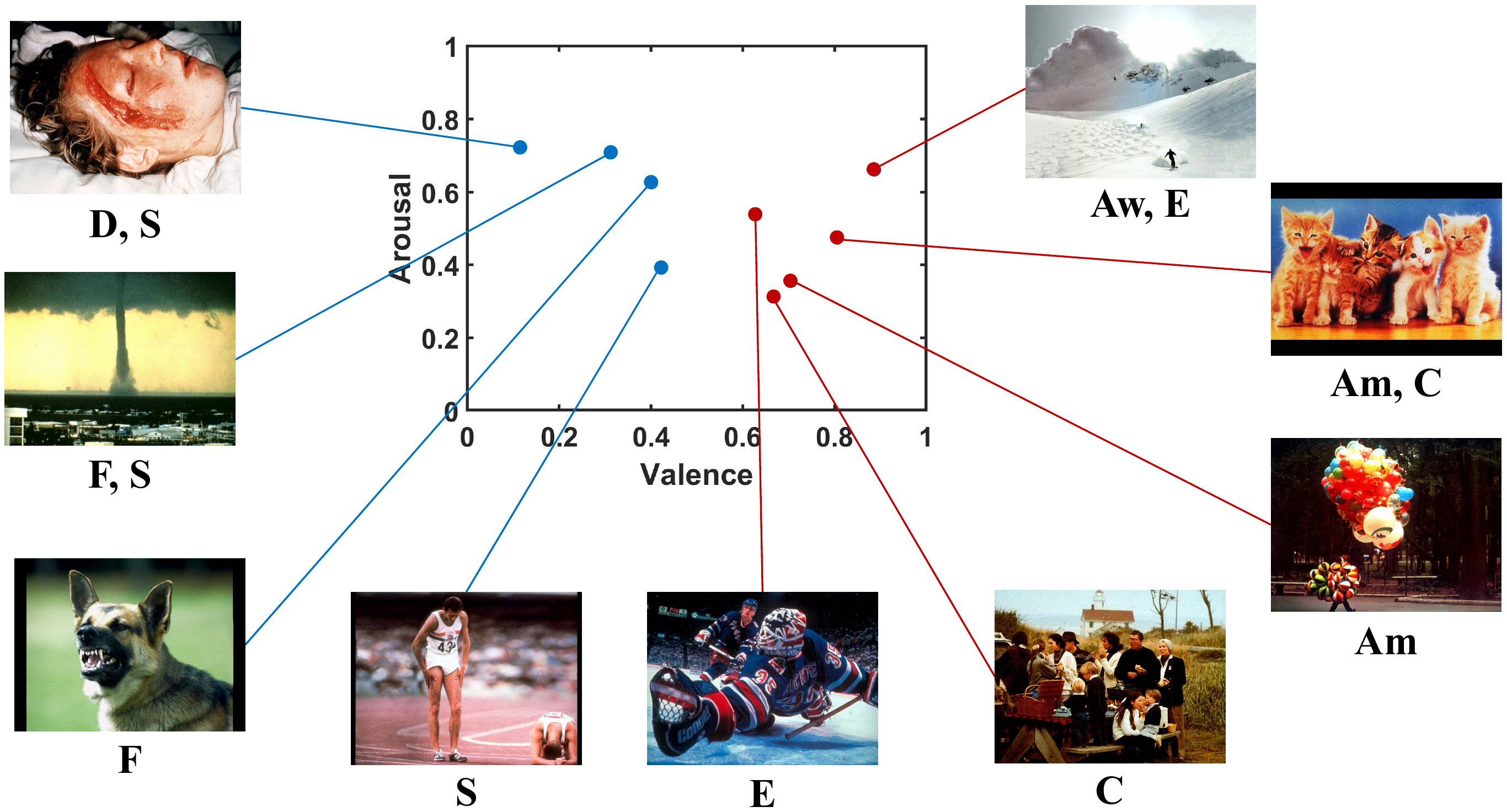}
\caption{Illustration of fine-grained continuous visual emotions v.s. coarse-grained discrete visual emotions from the IAPS dataset~\cite{lang1997international,mikels2005emotional}, where `Am', `Aw', `C', `D', `E', `F', and `S' are short for `amusement', `awe', `contentment', `disgust', `excitement', `fear', and `sadness', respectively. For simplicity, we only show the valence-arousal (VA) dimensions. It is clear that (1) the images belonging to the same discrete emotion category (\textit{e.g.} sadness) may have diverse VA values; (2) the same image that has unique VA value may correspond to different emotion categories. These observations motivate us to study the fine-grained visual emotion regression.}
\label{fig:Subtlety}
\end{center}
\end{figure}

Visual content is often used in psychology to evoke emotions in human viewers~\cite{detenber1998roll}. Nowadays, with the wide popularity of mobile devices, humans have become used to recording their activities, sharing their experiences, and expressing their opinions using images and videos with text in social networks like Flickr and Weibo~\cite{zhao2018real}. Automatically discovering the implied emotions from the huge volume of multimedia content can help in understanding humans' behaviors and preferences, which are crucial in many practical applications, such as smart advertising~\cite{yang2013integrated}, multimedia retrieval~\cite{zhao2014affective,yang2018retrieving}, and political voting forecasts~\cite{tumasjan2010predicting,chung2011can}.

Visual emotion analysis (VEA) is a high-level abstraction task, which aims to recognize the emotions induced by visual content. Because of the presence of two main challenges, \textit{i.e.} affective gap~\cite{hanjalic2006extracting,zhao2014exploring} between low-level visual features and high-level emotion semantics and the perception subjectivity~\cite{peng2015mixed,zhao2016predicting} among different viewers, VEA is a non-trivial problem. To bridge the affective gap, the key is to extract discriminative features~\cite{machajdik2010affective,lu2012shape,zhao2014exploring,rao2019learning,you2016building}. To tackle the subjectivity issue, we can predict personalized emotion perceptions for each viewer~\cite{zhao2018predicting}, or learn the emotion distributions for each image~\cite{peng2015mixed,yang2017joint,zhao2017approximating,zhao2017continuous,zhao2018discrete}. With the advent of deep learning, the research emphasis on VEA has shifted from traditional hand-crafted features designing~\cite{machajdik2010affective,lu2012shape,zhao2014exploring} to end-to-end deep representation learning with convolutional neural networks (CNNs)~\cite{peng2015mixed,rao2019learning,you2016building,yang2017joint,you2017visual,zhu2017dependency,song2018boosting,yang2018weakly}. These CNN-based methods either focus on dominant emotion classification~\cite{rao2019learning,you2016building,you2017visual,zhu2017dependency,song2018boosting,yang2018weakly,zhao2019cycleemotiongan} or emotion distribution learning~\cite{peng2015mixed,yang2017joint,zhao2018emotiongan}. Meanwhile, due to the similar mechanism to the human visual system and ability to model context information, visual attention has recently been incorporated into CNNs for visual emotion classification~\cite{you2017visual,song2018boosting,yang2018weakly} and achieves state-of-art performances.

However, there are several issues with above-mentioned CNN methods on the affective gap challenge:

First, these methods mainly focus on coarse-grained visual emotion classification, \textit{i.e.} assigning a dominant emotion category to an image, based on the emotion model of discrete emotion states (DES), such as the Mikels' eight emotions~\cite{mikels2005emotional} and binary sentiment. As emotions are naturally \emph{complex}, \emph{subjective}, and \emph{subtle}~\cite{peng2015mixed,zhao2016predicting,rao2019learning,yang2018weakly}, it is obviously insufficient to model emotion at such a coarse-grained level. As shown in Figure~\ref{fig:Subtlety}, the images belonging to the same emotion category may greatly differ in the continuous valence-arousal-dominance (VAD) space~\cite{schlosberg1954three}. On the other hand, discrete emotions do not provide a one-to-one relationship between the visual content and emotions~\cite{mikels2005emotional}, which highlights the utility of a dimensional approach.

Second, the attention based CNN methods on VEA only consider spatial attention. Although spatial attention modulates the local spatial connectivity patterns along each channel via spatially attentive weights~\cite{you2017visual,song2018boosting,yang2018weakly}, it neglects the interdependency between different channels. Nevertheless, channel-wise attention is very important, which can be viewed as a process of selecting semantic attributes and is essentially consistent with the CNN features~\cite{chen2017sca}.

In this paper, we study the fine-grained visual emotion regression problem to enrich the descriptive power of emotions based on VAD dimensions. The dimensional emotion values not only control for the inter-correlated nature of human emotions evoked by images~\cite{lu2012shape}, but also is more consistent with how the brain is organized to process emotions at their most basic level~\cite{lang1998emotion,lindquist2012brain}. Specifically, we design a novel network architecture, \textit{i.e.} Polarity-consistent Deep Attention Network (PDANet), that integrates attention into a CNN with an emotion polarity constraint for fine-grained visual emotion regression. First, both spatial and channel-wise attentions are incorporated into a CNN with mean squared error (MSE) loss. In this way, both the local spatial context along each channel and the interdependency between different channels are taken into account. Second, according to the assumption that VAD dimensions can be classified into different polarities~\cite{russell1999bipolarity,zhang2016exploring,subramanian2018ascertain,mollahosseini2019affectnet,zhao2019personalized}, we propose a novel polarity-consistent regression (PCR) loss based on the weakly supervised emotion polarity. The penalty of the predictions that have opposite polarity to the ground truth is increased. In this way, the polarity can be viewed as a constraint to guide the attention generation. With the polarity preserved attention map, the proposed PDANet method can obtain better visual emotion regression performances.

In summary, the contributions of this paper are threefold:

\begin{enumerate}
\item We propose to study the fine-grained visual emotion regression problem based on deep learning techniques. To the best of our knowledge, this is the first deep method for regressing visual emotions.
\item We develop a novel network architecture, \textit{i.e.} PDANet, that integrates both spatial and channel-wise attentions into a CNN with an emotion polarity constraint for visual emotion regression. Based on the weakly supervised emotion polarity, we propose a novel PCR loss, which enables PDANet to generate polarity preserved attention map and thus improve the emotion regression results.
\item We conduct extensive experiments on the IAPS~\cite{lang1997international}, NAPS~\cite{marchewka2014nencki}, and EMOTIC~\cite{kosti2017emotion} datasets, and the results demonstrate the superiority of the proposed PDANet method, as compared to the state-of-the-art approaches. The summarized datasets, baselines, evaluation metrics, and the reported results provide a systematic benchmark for future research on the visual emotion regression task.
\end{enumerate}

\section{Related Work}
\label{sec:RelatedWork}

\subsection{Visual Emotion Analysis}
\label{ssec:VEA}

Most existing methods on visual emotion analysis (VEA) either design hand-crafted features or employ deep learning frameworks to bridge the ``affective gap''~\cite{hanjalic2006extracting}. In the early years, numerous hand-crafted features were designed at different levels for VEA, such as low-level ones like Wiccest and Gabor~\cite{yanulevskaya2008emotional}, color~\cite{machajdik2010affective,sartori2015s,alameda2016recognizing}, texture~\cite{machajdik2010affective}, and shape~\cite{lu2012shape}; mid-level ones such as composition~\cite{machajdik2010affective}, sentributes~\cite{yuan2013sentribute}, principles-of-art~\cite{zhao2014exploring}, and bag-of-visual-words~\cite{rao2016multi}; and high-level ones such as adjective noun pairs (ANP)~\cite{borth2013large}. Some methods also fused different levels of features for affective image retrieval~\cite{zhao2014affective}, personalized emotion prediction~\cite{zhao2016predicting,zhao2018predicting}, and emotion distribution learning~\cite{zhao2017approximating,zhao2017learning,zhao2017continuous,zhao2018discrete}.

\begin{figure*}[!t]
\begin{center}
\centering \includegraphics[width=1.0\linewidth]{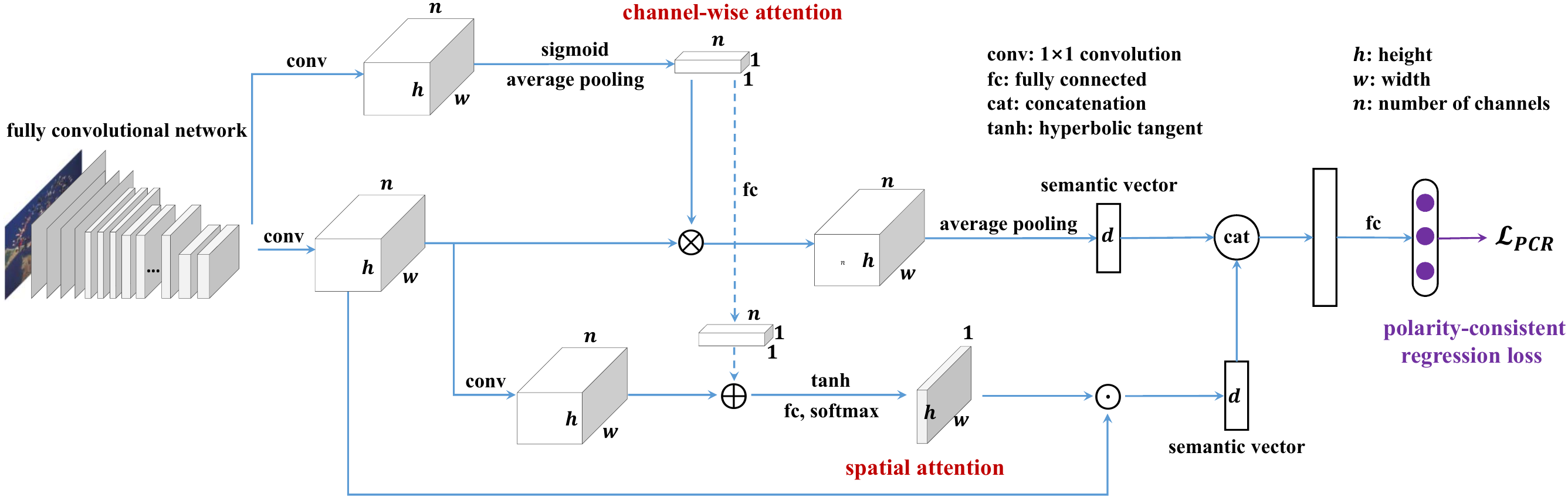}
\caption{Illustration of the proposed PDANet for fine-grained visual emotion regression. The input image is first fed into the convolutional layers of the fully convolutional network (FCN) ResNet-101. The response feature maps are then fed into two $1\times1$ convolutional layers; one is used to estimate channel-wise attention, and the other for spatial attention. The attended semantic vectors that capture the global and local information respectively are concatenated for emotion regression. Meanwhile, a novel polarity-consistent regression loss is optimized to guide the attention generation.}
\label{fig:Framework}
\end{center}
\end{figure*}

Recently, with the success of convolutional neural networks (CNNs) on many computer vision tasks, researchers have also applied CNNs to the VEA tasks. Extended from SentiBank~\cite{borth2013large}, DeepSentiBank~\cite{chen2014deepsentibank} constructs a visual sentiment concept to classify ANP for detecting emotions depicted in images. \citeauthor{peng2015mixed}~\shortcite{peng2015mixed} fine-tuned a pre-trained CNN to predict emotion distributions. A progressive CNN architecture is designed to make use of noisily labeled data for binary sentiment classification~\cite{you2016building}. \citeauthor{rao2019learning}~\shortcite{rao2019learning} proposed to learn multi-level deep representations for image emotion classification (MldrNet), which includes 3 different CNN models: Alexnet~\cite{krizhevsky2012imagenet}, aesthetics CNN, and texture CNN. On the basis of MldrNet, different levels of features are integrated with a bidirectional Gated Recurrent Unit model to exploit their dependencies~\cite{zhu2017dependency}. \citeauthor{yang2017joint}~\shortcite{yang2017joint} designed a multi-task deep framework by jointly optimizing emotion classification and distribution learning tasks. To explore the correlation of emotional labels with the same polarity, \citeauthor{yang2018retrieving}~\shortcite{yang2018retrieving} employed deep metric learning to optimize both the retrieval and classification tasks by jointly minimizing the traditional cross-entropy loss and a novel sentiment constraint.

Different from the above-mentioned methods, which try to improve the global image representations, several methods \cite{li2012context,yang2018visual,you2017visual,song2018boosting,yang2018weakly,rao2019multi} consider local information for VEA. While the regions in \cite{li2012context,yang2018visual,yang2018weakly} are produced by some segmentation or detection methods, such as EdgeBoxes~\cite{zitnick2014edge} and faster R-CNN~\cite{ren2015faster}, the local information is reflected by spatial attention in~\cite{you2017visual,song2018boosting,yang2018weakly,yao2019attention}. There are also some methods on domain adaptation~\cite{zhao2018emotiongan,zhao2019cycleemotiongan} and zero-shot learning~\cite{zhan2019zero} to deal with the label absence problem. Please refer to~\cite{joshi2011aesthetics,zhao2018affective} for a more comprehensive survey of VEA.

All of these CNN-based methods focus on coarse-grained visual emotion classification, \textit{i.e.} assigning a dominant discrete emotion category to an image. Differently, in this paper, we propose to study fine-grained visual emotion regression based on CNNs by incorporating visual attention and a polarity constraint. Although it is not easy for users to understand, the regression result is more powerful and can be used for more accurate emotion analysis during background processing.

\subsection{Visual Attention}
\label{ssec:VisualAttention}

Visual attention refers to a set of cognitive operations that allow us to efficiently deal with the limited processing capacity of the visual system by selecting relevant information and by filtering out irrelevant information~\cite{mcmains2009visual}. That is, we divert our attention to the regions of interest instead of the whole image. In this way, we can consider visual attention as a dynamic feature extraction mechanism that combines contextual fixations over time~\cite{mnih2014recurrent,stollenga2014deep,chen2017sca}.

As a widely studied topic in multimedia and computer vision, visual attention has recently been seamlessly incorporated into deep learning architectures and achieved outstanding performances in many vision-related tasks, such as image classification~\cite{larochelle2010learning,mnih2014recurrent,zhao2017diversified,hu2018squeeze,woo2018cbam}, image generation~\cite{tang2014learning}, image captioning~\cite{xu2015show,jia2015guiding,you2016image,chen2017sca,lu2017knowing,chen2018show,anderson2018bottom}, video captioning~\cite{gao2017video,song2017hierarchical}, video representation learning~\cite{zhao2017videowhisper}, and visual question answering~\cite{xu2016ask,yang2016stacked,lu2016hierarchical,anderson2018bottom}, in addition to non-vision related tasks, such as text sentiment analysis~\cite{chen2018twitter}. These attention methods can be roughly divided into three categories: spatial attention~\cite{xu2015show,xu2016ask,yang2016stacked}, semantic attention~\cite{jia2015guiding,you2016image,chen2018show}, and channel-wise attention~\cite{chen2017sca,hu2018squeeze}.

There are also several methods that employ attention for VEA~\cite{you2017visual,song2018boosting,yang2018weakly,yao2019attention}. \citeauthor{you2017visual}~\shortcite{you2017visual} directly estimated the spatial attention. \citeauthor{song2018boosting}~\shortcite{song2018boosting} employed a saliency map~\cite{li2016deep} as a prior knowledge and regularizer to holistically refine the attention distribution. \citeauthor{yang2018weakly}~\shortcite{yang2018weakly} proposed a weakly supervised coupled CNN with two branches: one is used to detect an emotion-specific soft map and the other combines the holistic and localized information for emotion classification. \citeauthor{yao2019attention}~\shortcite{yao2019attention} jointly modeled the polarity- and emotion-specific attended features. All these methods only consider spatial attention for visual emotion classification or retrieval. Differently from these, we propose a novel network architecture that integrates both spatial and channel-wise attention into a CNN for visual emotion regression. Further, we design a novel polarity-consistent regression loss to guide the attention generation.

\section{PDANet: Polarity-consistent Deep Attention Network}
\label{sec:PDANet}

An overview of the proposed polarity-consistent deep attention network (PDANet) is illustrated in Figure~\ref{fig:Framework}. The goal is to learn a discriminative model from images with attentive regions for fine-grained visual emotion regression where only image-level VAD labels are provided. Specifically, the proposed PDANet consists of three main components: a pre-trained FCN which is exploited to learn image representations, a two-branch multi-layer neural network to estimate both the spatial attention and channel-wise attention, and a fully connected layer to optimize the emotion regression task. The training of PDANet is performed by minimizing the polarity-consistent regression (PCR) loss. The whole framework is trained in an end-to-end manner.

\subsection{Image Representation Extraction}
\label{ssec:Representation}

Recent research has convincingly demonstrated that CNNs are highly capable of learning visual representations. Generally, the extracted features from different CNN layers usually have different properties~\cite{song2018boosting}. For example, the features at the bottom layers tend to reflect the low-level information (\textit{e.g.} texture), while the features at the top layers usually correspond to high-level semantics. Following~\cite{song2018boosting,yang2018weakly}, we use the output of fully convolutional layers to represent local regions of an image. Without loss of generality and following~\cite{yang2018weakly}, we choose a widely used CNN architecture, \textit{i.e.} ResNet-101~\cite{he2016deep}, as our basic CNN model to extract image representations. Suppose we are given $N$ training samples $\{(\textbf{x}_i,\textbf{y}_i)\}_{i=1}^{N}$, where $\textbf{x}_i$ is an affective image, $\textbf{y}_i=(v_i,a_i,d_i)$ is the corresponding continuous VAD value. For image $\textbf{x}_i$, suppose the feature maps of the conv5 in ResNet-101 are $\textbf{F}_i\in \mathbb{R}^{h\times w\times n}$ (we omit $i$ for simplicity in the following), where $h$ and $w$ are the spatial size (height and width) of the feature map, and $n$ is the number of channels. We reshape $\textbf{F}$ as
\begin{equation}
\textbf{F}=[\textbf{f}_1,\textbf{f}_2,\cdots,\textbf{f}_m]\in\mathbb{R}^{n\times m},
\end{equation}
by flattening the height and width of the original $\textbf{F}$, where $\textbf{f}_j\in\mathbb{R}^{n}$ and $m=h\times w$. Here we can consider $\textbf{f}_j$ as the visual feature of the $j$-th location.

\subsection{Spatial Attention Estimation}
\label{ssec:Spatial}
Image regions not only play an important role in image captioning~\cite{xu2015show,jia2015guiding,you2016image,chen2017sca,lu2017knowing,chen2018show,anderson2018bottom} and visual question answering~\cite{xu2016ask,yang2016stacked,lu2016hierarchical,anderson2018bottom}, but are also crucial in expressing emotions \cite{you2017visual,song2018boosting,yang2018weakly}. Take the picnic image with discrete label `C' in Figure~\ref{fig:Subtlety} for example. There are several objects in the image, including \emph{adult}, \emph{child}, \emph{table}, \emph{chair}, \emph{basket}, \emph{cabin}, \emph{etc}. Obviously, these objects contribute differently to predicting the emotions of this image. Therefore, directly applying a global visual feature to recognize emotion may lead to sub-optimal results due to the irrelevant regions (\textit{e.g.} \emph{table}, \emph{chair}, \emph{basket}, \emph{cabin}). In this paper, we employ the spatial attention mechanism to emphasize the emotional semantic-related regions. Following~\cite{chen2017sca,song2018boosting}, we employ a multi-layer neural network, \textit{i.e.} two $1\times 1$ convolutional layers and then a hyperbolic tangent function to generate the spatial attention distributions $\textbf{A}_S$ over all the image regions. That is
\begin{equation}
\begin{aligned}
&\textbf{H}_S=\textbf{W}_{S1}\tanh(\textbf{W}_{S2}\textbf{F}\oplus \textbf{b}_S),\\
&\textbf{A}_S=\text{Softmax}(\textbf{H}_S),\\
\end{aligned}
\label{equ:spatial}
\end{equation}
where $\textbf{W}_{S2}\in\mathbb{R}^{k\times n}$ and $\textbf{W}_{S1}\in\mathbb{R}^{1\times k}$ are two parameter matrices, $k$ is the size of hidden layers, $\textbf{b}_S\in\mathbb{R}^{k}$ is a $k$-dimensional bias vector, and $\oplus$ denotes the addition of a matrix and a vector, which is performed by adding the vector to each column of the matrix.

Then a $d$ dimensional semantic vector based on spatial attention is obtained as follows
\begin{equation}
\textbf{f}_S=\text{sp}(\textbf{A}_S\odot(\textbf{W}_{S2}\textbf{F}\oplus \textbf{b}_S)),
\end{equation}
where $\text{sp}$ is short for sum pooling, and $\odot$ is the multiplication of a matrix and a vector, which is performed by multiplying each value in the vector to each column of the matrix.

\subsection{Channel-wise Attention Estimation}
\label{ssec:ChannelWise}

Based on the assumption that each channel of a feature map in a CNN is a response activation of the corresponding convolutional layer, channel-wise attention can be viewed as a process of selecting semantic attributes~\cite{chen2017sca}. To generate the channel-wise attention, we first reshape $\textbf{F}$ to $\textbf{G}$
\begin{equation}
\textbf{G}=[\textbf{g}_1,\textbf{g}_2,\cdots,\textbf{g}_n]\in\mathbb{R}^{m\times n},
\end{equation}
where $\textbf{g}_j$ represents the $j$-th channel of the feature map $\textbf{F}$. The channel-wise attention $\textbf{A}_C$ is defined as
\begin{equation}
\textbf{A}_C=\text{ap}(\text{Sigmoid}(\textbf{W}_{C1}\textbf{G}\oplus \textbf{b}_C)),
\end{equation}
where $\textbf{W}_{C1}\in\mathbb{R}^{m\times m}$ is a transformation matrix, $\textbf{b}_C\in\mathbb{R}^{m}$ is a bias term, and $\text{ap}$ is short for average pooling. We can obtain another $d$ dimensional semantic vector based on channel-wise attention
\begin{equation}
\textbf{f}_C=\text{ap}(\textbf{A}_C\otimes(\textbf{W}_{S2}\textbf{F}\oplus \textbf{b}_S)),
\end{equation}
where $\otimes$ is the linear combination between a vector and a matrix, which is performed by multiplying each row of the matrix by the corresponding element of the vector.

Further, we also update the spatial attention based on the channel-wise attention
\begin{equation}
\begin{aligned}
&\textbf{H}_S=\textbf{W}_{S1}\tanh((\textbf{W}_{S2}\textbf{F}\oplus \textbf{b}_S)\oplus \text{fc}(\textbf{A}_C)),\\
&\textbf{A}_S=\text{Softmax}(\textbf{H}_S),\\
\end{aligned}
\label{equ:spatialNew}
\end{equation}
where $\text{fc}$ is short for fully connected.

\subsection{Polarity-consistent Regression Loss}
\label{ssec:PCR}
We concatenate $\textbf{f}_S$ and $\textbf{f}_C$ to obtain an aggregated semantic vector $\textbf{f}_{A}=[\textbf{f}_S^T,\textbf{f}_C^T]^T$, which can be viewed as the final visual representation and fed into a fully connected layer to predict the emotion labels. The mean squared error (MSE) loss of emotion regression is defined as
\begin{equation}
\mathcal{L}_{reg}=\frac{1}{N}\sum_{i=1}^{N}\sum_{j=1}^{N_E}(\text{fc}(\textbf{f}_{Ai})_j-\textbf{y}_{ij})^2,
\label{equ:regLoss}
\end{equation}
where $N_E$ is the dimension number of the adopted emotion model ($N_E=3$ in this paper), and $\textbf{y}_{ij}$ indicates the emotion label of the $j$-th dimension, \textit{e.g.} $\textbf{y}_{i1}=v_i$.

Directly optimizing the emotion regression loss in Eq.~(\ref{equ:regLoss}) may result in attended regions of the objects that have totally different emotion polarity from the visual emotion. Take the balloon image with label `Am' in Figure~\ref{fig:Subtlety} for example. The object \emph{chair} surrounded by dry grass and fallen leaves tends to convey negative emotion. In this paper, we proposed a polarity-consistent regression (PCR) loss to guide the attention generation based on the assumption that VAD dimensions can be classified into different polarities~\cite{russell1999bipolarity,zhang2016exploring,subramanian2018ascertain,mollahosseini2019affectnet,zhao2019personalized}. That is, the penalty of the predictions that have opposite polarity to the ground truth is increased. The PCR loss is defined as
\begin{equation}
\mathcal{L}_{PCR}=\frac{1}{N}\sum_{i=1}^{N}\sum_{j=1}^{N_E}(\text{fc}(\textbf{f}_{Ai})_j-\textbf{y}_{ij})^2(1+\lambda g(\text{fc}(\textbf{f}_{Ai})_j,\textbf{y}_{ij})),
\label{equ:Loss}
\end{equation}
where $\lambda$ is a penalty coefficient that controls the penalty extent. Similar to the indicator function, g(.,.) represents whether to add the penalty or not and is defined as
\begin{equation}
g(\hat{y},y)=\begin{cases}
1, & \text{if } p(\hat{y})\neq p(y),\\
0, & otherwise,
\end{cases}
\label{equ:indicator}
\end{equation}
where $p(.)$ is a function to compute the polarity of given dimensional emotions. In our experiment, $\lambda$ is selected on the validation set and $p(.)$ is set as a dichotomization function as in~\cite{subramanian2018ascertain}. Since the derivatives with respect to all parameters can be computed, we can train the proposed PDANet effectively in an end-to-end manner using stochastic gradient descent (SGD) to minimize the loss function in Eq.~(\ref{equ:Loss}).

\section{Experiments}
\label{sec:Experiments}

In this section, we first introduce the detailed experimental settings, including the datasets, baselines, evaluation metrics, and implementation details. We then evaluate the performance of the proposed method as compared to the state-of-the-art approaches. Finally, we conduct an ablation study to analyze the impact of different components and visualize the attention maps to further demonstrate the effectiveness of the proposed method.

\begin{table*}[!t]
\centering
\caption{Performance comparison measured by $MSE$ ($\times 10^{-2}$), where `FT', `V', `A', `D', and `M' are short for `Fine-tuned', `Valence', `Arousal', `Dominance', and `Mean', respectively. The best method is emphasized in bold. Our method achieves the best results on most metrics, significantly outperforming the state-of-the-art. Please see Section~\ref{ssec:Baselines} for more details of the baselines.}
\begin{tabular}
{c | c c c c | c c c c | c c c c}
\toprule
\multirow{2}{*}{Method} & \multicolumn{4}{c|}{IAPS} & \multicolumn{4}{c|}{NAPS} & \multicolumn{4}{c}{EMOTIC} \\
 & V & A & D & M & V & A & D & M & V & A & D & M \\
\hline
PAEF~\cite{zhao2014exploring}  &  5.431 & 2.050 & 1.893 & 3.125 & 4.353 & 1.723 & 3.482 & 3.186 & 2.015 & 4.211 & 3.230 & 3.152 \\
SentiBank~\cite{borth2013large}  & 5.383 & 2.018 & 1.904 & 3.102 & 4.439 & 1.788 & 3.525 & 3.251 & 1.920 & 3.613 & 3.084 & 2.872 \\
AlexNet~\cite{krizhevsky2012imagenet}  & 5.174 & 1.998 & 1.903 & 3.025 & 4.378 & 1.789 & 3.499 & 3.222 & 1.882 & 3.444 & \textbf{3.046} & 2.791 \\
VGG-16~\cite{simonyan2015very}  & 5.089 & 2.031 & 1.835 & 2.985 & 4.460 & 1.765 & 3.500 & 3.242 & 2.021 & 4.403 & 3.327 & 3.250 \\
ResNet-101~\cite{he2016deep}  & 3.456 & \textbf{1.266} & 1.222 & 1.981 & 2.577 & 1.061 & 2.002 & 1.880 & 1.866 & 3.269 & 3.061 & 2.732 \\
\hline
FT AlexNet~\cite{you2016building}  & 4.479 & 1.968 & 2.993 & 3.147 & 3.701 & 1.556 & 2.993 & 2.750 & 2.202 & 3.699 & 3.121 & 3.007 \\
FT VGG-16~\cite{you2016building}  & 3.560 & 1.672 & 1.450 & 2.227 & 3.096 & 1.130 & 2.335 & 2.187 & 1.895 & 3.414 & 3.069 & 2.793 \\
FT ResNet-101~\cite{you2016building}  & 3.214 & 1.597 & 1.844 & 2.218 & 2.252 & \textbf{0.955} & 1.844 & 1.684 & 1.844 & 3.304 & 3.093 & 2.747 \\
MldrNet~\cite{rao2019learning}  & 5.219 & 2.073 & 2.005 & 3.099 & 4.012 & 2.413 & 3.170 & 3.198 & 2.059 & 4.266 & 3.242 & 3.189 \\
\hline
SentiNet-A~\cite{song2018boosting}  & 3.952 & 1.975 & 1.707 & 2.545 & 2.943 & 1.440 & 2.430 & 2.271 & 1.902 & 3.445 & 3.078 & 2.808 \\
WSCNet~\cite{yang2018weakly}  & 3.255 & 1.571 & 1.524 & 2.117 & 2.951 & 1.181 & 2.266 & 2.133 & 1.849 & 3.367 & 3.153 & 2.790 \\
\textbf{PDANet (ours)}  & \textbf{3.179} & 1.279 & \textbf{1.221} & \textbf{1.893} & \textbf{2.248} & 0.971 & \textbf{1.793} & \textbf{1.671} & \textbf{1.776} & \textbf{3.261} & 3.076 & \textbf{2.704} \\
\bottomrule
\end{tabular}
\label{tab:MSE}
\end{table*}

\begin{table*}[!t]
\centering
\caption{Performance comparison measured by $R^2$ ($\times 10^{-1}$). The best method is emphasized in bold. The proposed PDANet significantly outperforms the baselines.}
\begin{tabular}
{c | c c c c | c c c c | c c c c}
\toprule
\multirow{2}{*}{Method} & \multicolumn{4}{c|}{IAPS} & \multicolumn{4}{c|}{NAPS} & \multicolumn{4}{c}{EMOTIC} \\
 & V & A & D & M & V & A & D & M & V & A & D & M \\
\hline
PAEF~\cite{zhao2014exploring}  & -0.491 & -0.454 & -0.384 & -0.443 & -0.261 & 0.143 & -0.222 & -0.113 & -0.006 & 0.337 & 0.037 & 0.123 \\
SentiBank~\cite{borth2013large}  &  -0.399 & -0.294 & -0.443 & -0.378 & -0.464 & -0.233 & -0.351 & -0.349 & 0.466 & 1.709 & 0.488 & 0.888 \\
AlexNet~\cite{krizhevsky2012imagenet}  & 0.005 & -0.191 & -0.436 & -0.207 & -0.321 & -0.234 & -0.273 & -0.276 & 0.654 & 2.097 & \textbf{0.604} & 1.118 \\
VGG-16~\cite{simonyan2015very}  &  0.170 & -0.355 & -0.061 & -0.082 & -0.513 & -0.099 & -0.275 & -0.296 & -0.038 & -0.104 & -0.261 & -0.135\\
ResNet-101~\cite{he2016deep}  & 3.324 & \textbf{3.546} & 3.297 & 3.389 & 3.926 & 3.931 & 4.121 & 3.993 & 0.733 & 2.499 & 0.558 & 1.263 \\
\hline
FT AlexNet~\cite{you2016building}  & 0.250 & 0.191 & -0.824 & -0.128 & 0.761 & -0.216 & -0.421 & 0.041 & 0.057 & 1.510 & 0.373 & 0.647 \\
FT VGG-16~\cite{you2016building}  &  2.250 & 1.667 & 1.571 & 1.829 & 2.270 & 2.582 & 1.871 & 2.241 & 0.587 & 2.165 & 0.533 & 1.095 \\
FT ResNet-101~\cite{you2016building}  & 3.003 & 2.038 & 1.429 & 2.157 & 3.058 & 3.728 & 2.144 & 2.977 & 0.840 & 2.416 & 0.459 & 1.238 \\
MldrNet~\cite{rao2019learning}  & -0.081 & -0.569 & -0.993 & -0.548 & 0.543 & -3.804 & 0.693 & -0.856 & -0.224 & 0.211 & -0.002 & -0.005 \\
\hline
SentiNet-A~\cite{song2018boosting}  & 2.365 & -0.071 & 0.638 & 0.978 & 3.062 & 1.759 & 2.866 & 2.562 & 0.553 & 0.209 & 0.505 & 0.422 \\
WSCNet~\cite{yang2018weakly}  & 2.915 & 2.170 & 1.040 & 2.042 & 2.632 & 2.247 & 2.110 & 2.330 & 0.816 & 2.272 & 0.272 & 1.120 \\
\textbf{PDANet (ours)}  &  \textbf{3.859} & 3.479 & \textbf{3.305} & \textbf{3.548} & \textbf{4.701} & \textbf{4.443} & \textbf{4.737} & \textbf{4.627} & \textbf{1.180} & \textbf{2.515} & 0.511 & \textbf{1.402} \\
\bottomrule
\end{tabular}
\label{tab:R2}
\end{table*}

\subsection{Datasets}
\label{ssec:Datasets}
To evaluate the performances of the proposed method, we employ 3 publicly available datasets that contain continuous emotion labels: IAPS \cite{lang1997international}, NAPS~\cite{marchewka2014nencki}, and EMOTIC~\cite{kosti2017emotion}.

The International Affective Picture System (IAPS)~\cite{lang1997international} is an emotion evoking image set in psychology. It consists of 1,182 documentary-style natural color images depicting complex scenes, such as \emph{portraits}, \emph{babies}, \emph{animals}, \emph{landscapes}, \emph{etc}. Each image is associated with an empirically derived mean and standard deviation (STD) of VAD ratings in a 9-point rating scale by about 100 college students.

The Nencki Affective Picture System (NAPS)~\cite{marchewka2014nencki} consists of 1,356 realistic, high-quality photographs with five categories, \textit{i.e.} \emph{people}, \emph{faces}, \emph{animals}, \emph{objects}, and \emph{landscapes}. These images were rated by 204 mostly European participants in a 9-point bipolar semantic sliding scale on the VA and approach-avoidance (we use dominance for simplicity) dimensions. On average, 55 ratings were collected for each image.

The Emotions in Context Database (EMOTIC)~\cite{kosti2017emotion} is a dataset of images containing people in context in non-controlled environments. It is composed of images from MSCOCO~\cite{lin2014microsoft}, Ade20k~\cite{zhou2019semantic}, and images downloaded from Google. The images were annotated by Amazon Mechanical Turk (AMT) workers with 26 emotion categories and the continuous 10-scale VAD dimensions. In total, there are 18,316 images in this dataset with 23,788 annotated people.

\subsection{Baselines}
\label{ssec:Baselines}
To the best of our knowledge, PDANet is the first work on deep learning based visual emotion regression. To demonstrate its effectiveness, we compare it with several baselines, including the methods using hand-crafted features, CNN-based methods, and attention based methods. \textbf{For the traditional methods}, we extract mid-level 165 dimensional principles-of-art based emotion features (PAEF)~\cite{zhao2014exploring}, and high-level 1,200 dimensional adjective noun pairs (ANP) with SentiBank~\cite{borth2013large}. We use support vector regression (SVR) with the radial basis function (RBF) kernel in the LIBSVM library~\cite{chang2011libsvm} as the regressor. \textbf{For the CNN-based methods}, following~\cite{you2016building}, we first fine-tune three classical deep learning methods pre-trained on ImageNet: AlexNet~\cite{krizhevsky2012imagenet}, VGG-16~\cite{simonyan2015very}, and ResNet-101~\cite{he2016deep}. Second, we show the results of fully connected features extracted from the ImageNet CNN, which are classified by SVR. We also compare PDANet with one specifically designed deep methods, \textit{i.e.} multi-level deep representations (MldrNet)~\cite{rao2019learning}. \textbf{For the attention-based methods}, we choose two recently published methods: SentiNet-A~\cite{song2018boosting} and WSCNet~\cite{yang2018weakly}, which respectively estimate the spatial attention with the saliency map as a prior regularizer and generate a sentiment map as a detection task in a weakly supervised manner. Please note that for both CNN and attention based methods, we replace the cross-entropy loss with MSE loss for the visual emotion regression task.

\subsection{Evaluation Metrics}
\label{ssec:Metrics}
We employ mean squared error ($MSE$) and R squared ($R^2$) to evaluate the visual emotion regression results. $MSE$ is a quadratic scoring rule that measures the average magnitude of the error. It is defined as the average of squared differences between the prediction and ground truth
\[
MSE = \frac{1}{M}\sum_{i=1}^{M}(z_i-\hat{z_i})^2,
\]
where $M$ is the number of testing samples, and $z_i$ and $\hat{z_i}$ are the ground truth and prediction of the VAD emotions, respectively. $MSE\geq 0$ and the smaller the better.

$R^2$, also known as the coefficient of determination, shows how well the predicted results explain the variability in the ground truth values. It is defined as
\[
R^2 = 1-\frac{\frac{1}{M}\sum_{i=1}^{M}(z_i-\hat{z_i})^2}{\frac{1}{M}\sum_{i=1}^{M}(z_i-\bar{z})^2},
\]
where $\bar{z}=\frac{1}{M}\sum_{i=1}^{M}z_i$ is the mean of the ground truth emotions. The numerator is MSE and the denominator is the variance of emotions. $R^2\leq 1$, with a larger value representing a better result. Please note that other regression measures, such as mean absolute error and root mean squared error, can also be used as evaluation metrics. Due to the page limit, we do not report these results.

\subsection{Implementation Details}
\label{ssec:Details}

Our model is based on the state-of-the-art CNN architecture ResNet-101~\cite{he2016deep}. The network is initialized with the weights from the pre-trained ResNet-101 model on ImageNet~\cite{deng2009imagenet}. In addition, we resize the image to $600\times600$ pixels, apply random horizontal flips, and crop a random $448\times448$ patch as a form of data augmentation to reduce overfitting. We replace the last layers (global average pooling and fully connected layer) with the proposed attention networks. We use a weight decay of 0.0005 with a momentum of 0.9, a batch size of 32, and fine-tune all layers with SGD. The learning rates of the convolutional layers and the last fully-connected layer are initialized as 0.001 and 0.01, respectively. For the IAPS and NAPS datasets, the total number of epochs is 300 with learning rate dropped by a factor of 10 for the last 50 epochs. For the EMOTIC dataset, the total number of epochs is 50 and the learning rate is dropped by a factor of 10 for the last 10 epochs. Since the EMOTIC dataset is initially constructed to recognize  emotions in context for the people contained in an image, there might be more than one VAD label for each image~\cite{kosti2017emotion}. We employ the average VAD values of all the people's emotions as the ground truth for an image in our emotion regression task. For all the three datasets, we randomly split each into 70\% training, 10\% validation, and 20\% testing. The VAD labels are normalized to [0, 1] for better comparison. At test time, we adopt the standard 10-crop testing and our prediction takes the regressors' output for final evaluation. Our model is implemented using PyTorch. All of our experiments are performed on 2 NVIDIA RTX 2080Ti GPUs with 11 GB memory.

\subsection{Comparison with the State-of-the-art}
\label{ssec:Comparison}

\begin{table*}[!t]
\centering
\caption{Ablation study of different components in the proposed PDANet for fine-grained visual emotion regression measured by $MSE$ ($\times 10^{-2}$), where `S', `CW', and `PCR' are short for spatial attention, channel-wise attention, and polarity-consistent regression loss, respectively. All the components contribute to the emotion regression task.}
\begin{tabular}
{c | c c c c | c c c c | c c c c}
\toprule
\multirow{2}{*}{Method} & \multicolumn{4}{c|}{IAPS} & \multicolumn{4}{c|}{NAPS} & \multicolumn{4}{c}{EMOTIC} \\
 & V & A & D & M & V & A & D & M & V & A & D & M \\
\hline
S        & 4.947 & 2.012 & 1.785 & 2.915 & 3.587 & 1.641 & 2.836 & 2.688 & 1.998 & 3.730 & 3.112 & 2.947 \\
CW       & 3.362 & 1.267 & 1.316 & 1.982 & 2.340 & 1.011 & 1.909 & 1.753 & 1.819 & 3.271 & 3.088 & 2.726 \\
S+CW     & 3.320 & 1.319 & 1.259 & 1.966 & 2.254 & 0.985 & 1.800 & 1.680 & 1.819 & 3.269 & 3.072 & 2.720 \\
S+CW+PCR & 3.179 & 1.279 & 1.221 & 1.893 & 2.248 & 0.971 & 1.793 & 1.671 & 1.776 & 3.261 & 3.076 & 2.704 \\
\bottomrule
\end{tabular}
\label{tab:MSEAblation}
\end{table*}

\begin{table*}[!t]
\centering
\caption{Ablation study of different components in the proposed PDANet for fine-grained visual emotion regression measured by $R^2$ ($\times 10^{-1}$). All the components contribute to the emotion regression task.}
\begin{tabular}
{c | c c c c | c c c c | c c c c}
\toprule
\multirow{2}{*}{Method} & \multicolumn{4}{c|}{IAPS} & \multicolumn{4}{c|}{NAPS} & \multicolumn{4}{c}{EMOTIC} \\
 & V & A & D & M & V & A & D & M & V & A & D & M \\
\hline
S        & 0.444 & -0.263 & 0.211 & 0.131 & 1.545 & 0.613 & 1.674 & 1.277 & 0.017 & 1.438 & 0.304 & 0.586 \\
CW       & 3.506 & 3.539  & 2.782 & 3.276 & 4.483 & 4.217 & 4.397 & 4.366 & 0.968 & 2.493 & 0.473 & 1.311 \\
S+CW     & 3.587 & 3.327  & 3.097 & 3.337 & 4.681 & 4.347 & 4.714 & 4.581 & 0.966 & 2.497 & 0.521 & 1.328 \\
S+CW+PCR & 3.859 & 3.479  & 3.305 & 3.548 & 4.701 & 4.443 & 4.737 & 4.627 & 1.180 & 2.515 & 0.511 & 1.402 \\
\bottomrule
\end{tabular}
\label{tab:R2Ablation}
\end{table*}

The performance comparisons between the proposed PDANet model and the state-of-the-art approaches as measured by $MSE$ and $R^2$ are shown in Table~\ref{tab:MSE} and Table~\ref{tab:R2}, respectively. From these results, we can observe that:

\textbf{(1)} The traditional methods, \textit{i.e.} those which extract visual features (either hand-crafted or CNN-based) and feed them to typical regressors, do not perform well. Due to the presence of \emph{affective gap}, the directly extracted features are likely to be inconsistent with the highly abstract emotions. One interesting observation is that the hand-crafted features perform comparably or even better than the pre-trained CNN features (\textit{e.g.} AlexNet, VGG-16). This shows the difference between emotion analysis and other computer vision problems. Images with similar appearances may convey totally different emotions, while the images containing quite different content may evoke the same emotions. The pre-trained CNN that is designed for objective vision tasks (\textit{e.g.} object classification) cannot well capture this phenomenon, while the artistic principles and semantic ANP features are more robust. As the adopted three datasets are mainly natural scenes, containing obvious semantics, ANP achieves better performance than PAEF, which is more suitable to deal with abstract and artistic images.

\textbf{(2)} In most cases, the fine-tuned CNN on the affective datasets outperform the corresponding methods in traditional learning paradigm, \textit{i.e.} those with separate feature extraction and regressor learning. The fine-tuned CNN has the capability to analyze visual emotions, as after fine-tuning, the CNN can learn to adapt to the emotion datasets. These results show the generalizability of fine-tuned CNNs on visual emotion regression.

\textbf{(3)} Generally, the attention based methods performs comparably to the traditional methods and CNN-based methods, which demonstrates the effectiveness of attention in visual emotion regression. Even though only spatial attention is considered (\textit{e.g.} WSCNet~\cite{yang2018weakly}), the performance is still improved. This is reasonable, because it can distill the spatially informative regions in an image and thus mitigate the influence of background and irrelevant objects.

\begin{figure*}[!t]
\begin{center}
\centering \includegraphics[width=0.99\linewidth]{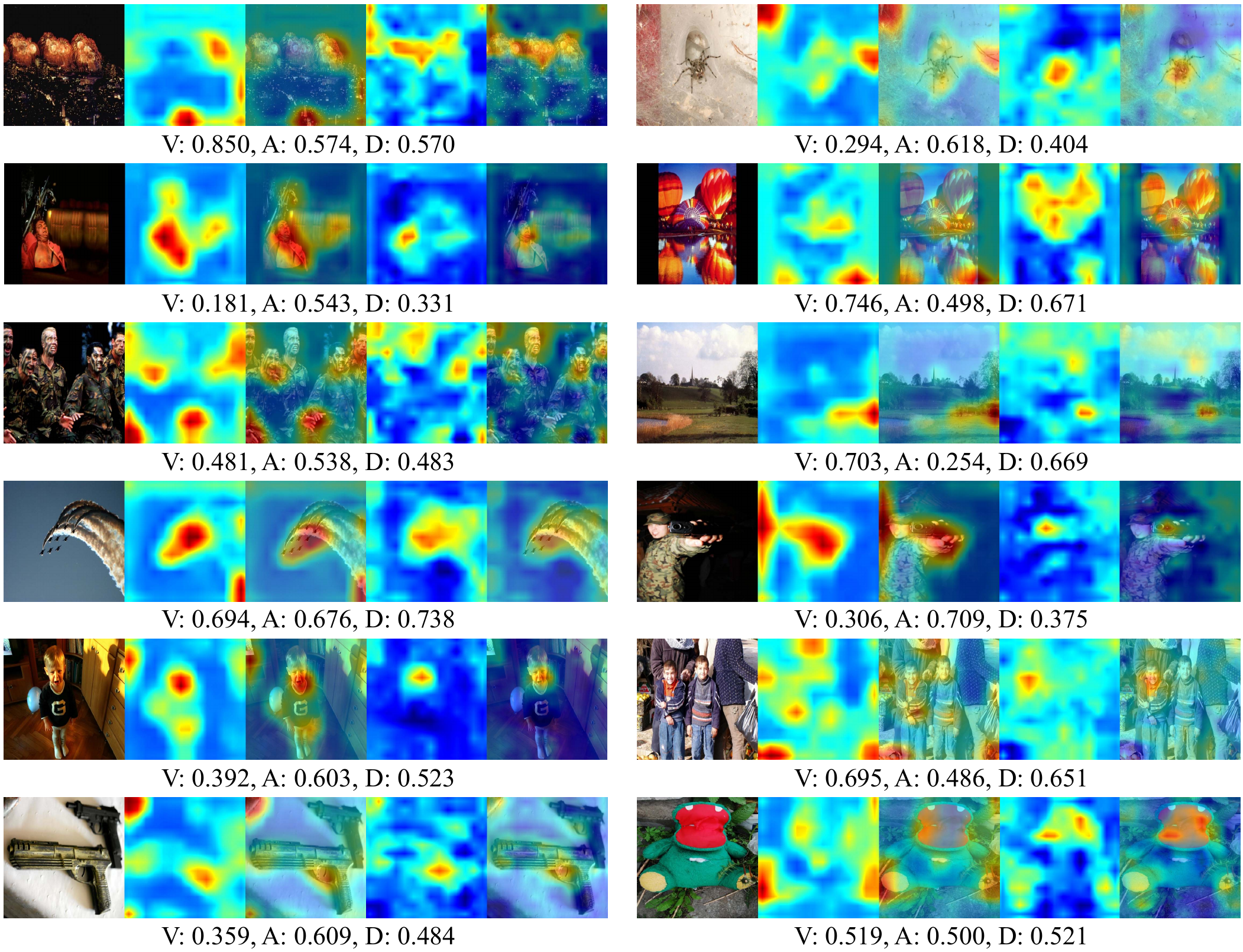}
\caption{Visualization of the learned attention maps. From left to right in each group are: original image from the test set, heat map by WSCNet~\cite{yang2018weakly}, the combination of image and heat map by WSCNet~\cite{yang2018weakly}, heat map by PDANet, and the combination by PDANet. The ground truth VAD values of the tested images are shown below each group. Red regions indicate more attention. The attention map of PDANet focuses more on the salient and discriminative regions for visual emotion regression.}
\label{fig:Visualization}
\end{center}
\end{figure*}

\textbf{(4)} The proposed PDANet model performs the best in almost all cases. We take the IAPS dataset and the mean result for example to quantitatively show the performance improvements. On the one hand, the relative performance improvements of PDANet over the traditional methods and CNN-based methods are 39.42\%, 38.98\%, 37.43\%, 36.58\%, 4.46\%, and 39.84\%, 15.01\%, 14.67\%, 38.92\%, respectively. On the other hand, the proposed PDANet achieves 25.61\% and 10.57\% relative performance improvements as compared to SentiNet-A~\cite{song2018boosting} and WSCNet~\cite{yang2018weakly}, respectively.

\textbf{(5)} These results demonstrate that the proposed PDANet model can achieve superior performance relative to the state-of-the-art approaches. The performance improvements benefit from the advantages of PDANet. First, the joint consideration of spatial attention and channel-wise attention takes into account the attentive information in context more thoroughly. Not only the importance of local spatial context along each channel, but also weights of the interdependency between different channels, are evaluated. Second, the PCR loss can guide the process of attention generation. With the polarity constraint, the generated attention is more likely to emphasize the regions that have the same polarity as the whole image and ignore the ones that are confusing for emotion regression.

\subsection{Ablation Study}
\label{ssec:Ablation}

The proposed PDANet model contains two major novel components: an attention generation strategy for incorporating both spatial and channel-wise context and a polarity-consistent regression (PCR) loss for guiding the attention generation. We conduct an ablation study to further verify their effectiveness. We begin with an experiment on spatial or channel-wise attention with traditional MSE regression loss. And then we test the performance of combining the spatial and channel-wise attentions. Finally, we add the PCR loss. The emotion regression results measured by $MSE$ and $R^2$ are shown in Table~\ref{tab:MSEAblation} and Table~\ref{tab:R2Ablation}, respectively.

From the results, we have the following observations: \textbf{(1)} Simply considering the spatial or channel-wise attention performs the worst with channel-wise attention slightly better than spatial, which shows the significance of the interdependency between different channels. \textbf{(2)} The combination of spatial and channel-wise attentions performs better than each single attention. This demonstrates the necessity of considering both spatial and channel-wise aspects when modeling attention for visual emotion regression. \textbf{(3)} All the components, \textit{i.e.} spatial attention, channel-wise attention, and the PCR loss, contribute to the visual emotion regression task. The proposed PDANet model that jointly combines the novel attention strategy and employs the PCR loss performs the best. These observations demonstrate the effectiveness of the proposed PDANet model.

\subsection{Visualization}
\label{ssec:Visualization}

In order to show the interpretability of our model,  we use the heat map generated by the Grad-Cam algorithm~\cite{gradcam2017iccv} to visualize the attention learned by WSCNet~\cite{yang2018weakly} and the proposed PDANet. As illustrated in Figure~\ref{fig:Visualization}, we observe that the attention maps generated by the proposed PDANet, can focus comparably or even better on the attentive and discriminative regions than WSCNet~\cite{yang2018weakly}, which employs a specific attention map detection branch. Take the group on the top left corner for example, PDANet focuses on the fireworks, which obviously determines the positive emotion, while WSCNet mainly concentrates on the background.

\section{Conclusion}
\label{sec:Conclusion}

In this paper, we make significant progress toward solving the fine-grained visual emotion regression problem with deep learning techniques. A novel network architecture, termed Polarity-consistent Deep Attention Network (PDANet), is developed by integrating visual attention into a CNN with a novel polarity-consistent regression (PCR) loss. Both spatial and channel-wise attentions are considered to model the local spatial context along each channel and the interdependency between different channels. The optimization of the PCR loss enables PDANet to generate polarity preserved attention maps, which can boost the emotion regression performance. The extensive experiments conducted on the IAPS, NAPS, and EMOTIC datasets demonstrate that PDANet significantly outperforms the state-of-the-art approaches for visual emotion regression. We also provided a systematic benchmark, including datasets, baselines, evaluation metrics, and results, for future research on visual emotion regression.

For further studies, we plan to extend the proposed PDANet model to other visual emotion analysis tasks, such as classification and retrieval. In addition, we will explore temporal attention in PDANet in order to attend features for emotion analysis of video content. Adapting a visual emotion regression model from a labeled source domain to another unlabeled target domain is also an interesting direction.

%
\begin{acks}
This work was supported by Berkeley DeepDrive, the National Natural Science Foundation of China (Nos. 61701273, 61571269), Natural Science Foundation of Jiangsu Province (No. BK20181354), and the National Key R\&D Program of China (No. 2017YFC011300).
\end{acks}

%
\small\bibliographystyle{ACM-Reference-Format}
\balance\bibliography{sample-base}

%

\end{document}